**A New Method for Vehicle Logo Recognition Based on Swin Transformer**

**Yang Li**
School of Optical-Electrical and Computer Engineering
University of Shanghai for Science and Technology, Shanghai, China, 200093
Email: 223330831@st.usst.edu.cn

**Doudou Zhang**
Principal Data Scientist
Capital One
Mclean, VA, USA, 22102
Email: Doudouzhang181@gmail.com

**Jianli Xiao (Corresponding Author)**
School of Optical-Electrical and Computer Engineering
University of Shanghai for Science and Technology, Shanghai, China, 200093
Email: audyxiao@sjtu.edu.cn

Word Count: 5704 words + 2 table (250 words per table) = 6,204 words

**ABSTRACT**

Intelligent Transportation Systems (ITS) utilize sensors, cameras, and big data analysis to monitor real-time traffic conditions, aiming to improve traffic efficiency and safety. Accurate vehicle recognition is crucial in this process, and Vehicle Logo Recognition (VLR) stands as a key method. VLR enables effective management and monitoring by distinguishing vehicles on the road. Convolutional Neural Networks (CNNs) have made impressive strides in VLR research. However, achieving higher performance demands significant time and computational resources for training. Recently, the rise of Transformer models has brought new opportunities to VLR. Swin Transformer, with its efficient computation and global feature modeling capabilities, outperforms CNNs under challenging conditions. In this paper, we implement real-time VLR using Swin Transformer and fine-tune it for optimal performance. Extensive experiments conducted on three public vehicle logo datasets (HFUT-VL1, XMU, CTGU-VLD) demonstrate impressive top accuracy results of 99.28%, 100%, and 99.17%, respectively. Additionally, the use of a transfer learning strategy enables our method to be on par with state-of-the-art VLR methods. These findings affirm the superiority of our approach over existing methods. Future research can explore and optimize the application of the Swin Transformer in other vehicle vision recognition tasks to drive advancements in ITS.

**INTRODUCTION**

Intelligent Transportation Systems (ITS) utilize advanced technologies such as sensors, cameras, and big data analysis to monitor and manage traffic conditions in real-time (*1*). Its objective is to improve traffic efficiency, reduce congestion, and enhance traffic safety. As an important branch of ITS, vehicle recognition technology plays a key role in effectively monitoring and managing vehicles. During the tracking and classification process of moving vehicles, it is typically necessary to identify the vehicles' identity information, including license plates, vehicle models, colors, and logos. Traditional license plate recognition technology has been widely used for vehicle identification in the field of transportation and has achieved certain success (*2*). However, with the increase of unlicensed and licensed vehicles, relying solely on license plate recognition has become difficult to accurately determine the true identity information of vehicles. As the identity element of the vehicle, the vehicle logo can reflect the basic information of the vehicle manufacturer, which is more distinctive and reliable than the license plate, vehicle model and vehicle color. Vehicle logo recognition (VLR) is capable of quickly and accurately identifying vehicles of different brands in complex traffic environments, providing a more comprehensive and reliable means of vehicle recognition for ITS. Therefore, VLR has gradually gained widespread attention in recent years.

In previous studies, many researchers focused on using classical feature extraction methods for VLR. However, these methods suffer from laborious feature extraction, sensitivity to environmental changes, and lack of self-learning capability to enhance generalization and robustness. To address these issues, many researchers have turned to deep learning methods and achieved impressive results. Convolutional Neural Networks (CNNs) have gained significant attention in the field of VLR due to their characteristics of spatial hierarchical representation, translation invariance, and automatic feature extraction. With the continuous advancement of technology, CNNs have been extensively optimized in terms of network depth and structure, leading to the development of a series of excellent models such as ResNet (*3*), MobileNet (*4*), ShuffleNet (*5*), and EfficientNet (*6*), among others.

However, most CNN-based vehicle logo recognition studies still have weaknesses in global modeling (*7*). Transformer has a unique self-attentive computation that enables global characterization through efficient parallel computation (*8*). Initially applied to Natural Language Processing, Transformers later found their way into the Computer Vision domain with the emergence of Vision Transformer (ViT), promoting advancements in multimodal fields (*9*). Nonetheless, ViT's global attention mechanism makes computations more complex (*10*). Swin Transformer, on the other hand, offers a more efficient modeling approach and has recently excelled in the field of Computer Vision (*11–13*).

In this paper, we adopt the method of fine-tuning Swin Transformer and perform VLR based on its following features: (1) Swin Transformer adopts a hierarchical construction method similar to CNNs, effectively reducing model complexity. (2) Swin Transformer utilizes sliding window techniques, enabling information interaction between different windows in a sliding manner. The rest of this paper is organized as follows: Section II briefly reviews related research on VLR. In Section III, we provide a detailed description of the Swin Transformer method. In Section IV, we report and discuss experimental results. Finally, we summarize our findings in

Section V.

**RELATED WORKS**

VLR methods can be divided into two categories: classical extraction of features methods and deep learning methods, based on whether manual intervention is required in the feature extraction process. In this section, we will focus on reviewing these two categories of VLR methods.

**VLR Methods Based on Classical Extraction of Features**

Scale-Invariant Feature Transform (SIFT), Histogram of Oriented Gradients (HOG), and Invariant Moments are methods that can extract distinctive image features and have been applied to the task of logo recognition. Psyllos et al. (*14*) developed an enhanced scale-invariant feature transform for logo recognition by performing feature matching and hypothesis verification on key points. Yu et al. (*15*) first extracted dense SIFT features and then quantified logo features into visual words, combining them with SVM for logo classification. Yu et al. (*16*) divided vehicle logos into overlapping blocks, then extracted improved POEM features and combined them with the CRC classifier for recognition. Lu et al. (*17*) fused HOG and Local Binary Pattern (LBP) features, utilizing a multi-scale spatial representation for VLR. Yu et al. (*18*) introduced a multi-layer pyramid network and a multi-codebook encoding method for fast and efficient logo recognition.

**VLR Methods Based on Deep Learning**

Deep learning methods have the advantage of autonomously extracting features without the need for manual intervention. These methods, trained models, exhibit stronger generalization and robustness, making them well-suited for real-world VLR. Huang et al. (*19*) proposed a pre-training strategy based on PCA to effectively reduce the computational cost of convolutional kernels in CNNs, making it suitable for real-time VLR. Soon et al. (*20*) employed deep CNNs to automatically learn and extract advanced features of vehicle logos. Chen et al. (*21*) proposed a joint framework based on capsule networks, capable of accurately recognizing vehicle logos even in complex environments. Yu et al. (*22*) introduced a deep neural network composed of a region proposal network and a cascaded convolutional capsule network, enabling direct recognition of vehicle logos in images without prior coarse localization. Wang et al. (*23*) utilized the YOLOv5 network and transfer learning to recognize vehicle logos. Agarwal et al. (*24*) directly applied the ResNet152v2 network to VLR, achieving good results. Amirkhani and Barshooi in (*25*) proposed a novel approach that extracts different regions of interest (ROI) from input images and trains unique networks with preprocessing blocks for each ROI. They achieved recognition by integrating and sharing knowledge through a blackboard system. Shi et al. (*26*) utilized the lightweight YOLOv5s network and employed the cross-entropy loss method for VLR.

**METHODS**

The overall architecture of our model is illustrated in **Figure 1**. Firstly, the input RGB

vehicle logo image is segmented into non-overlapping patches, where each patch can be considered as a "token". Secondly, the original channel dimension of these tokens is projected to an arbitrary dimension through a linear embedding layer. Next, several Swin Transformer blocks along with a Patch Merging module are applied to these tokens. Finally, the feature vectors are mapped to an output vector of the same dimension as the number of categories through a linear layer, which is used for the classification of vehicle logos.

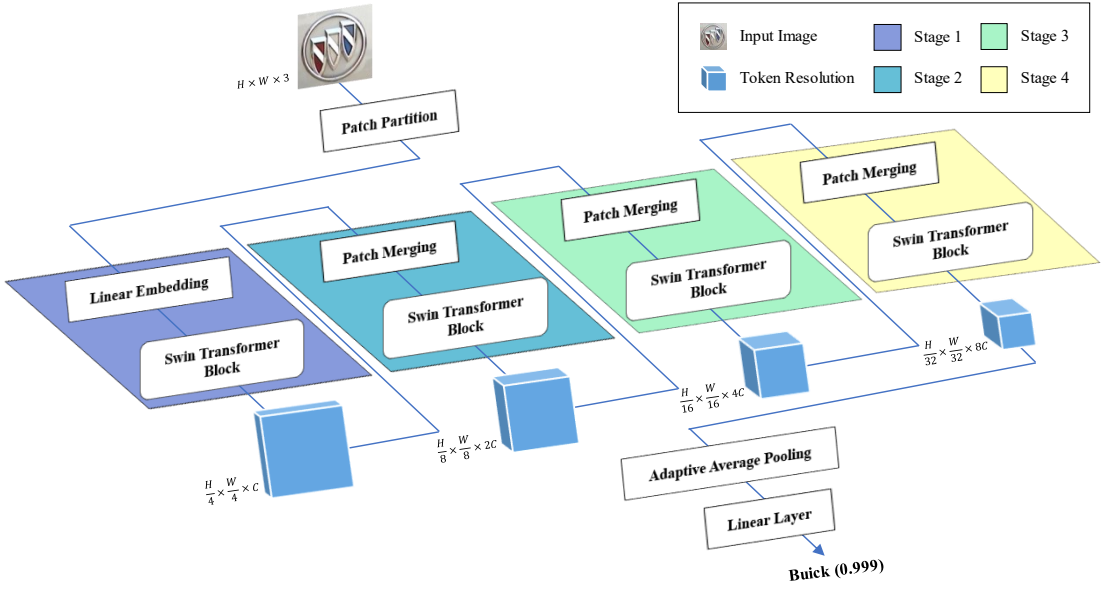

**Figure 1 The overall architecture of the Swin Transformer model**

**Swin Transformer model**

Swin Transformer, as a novel visual model following ViT, shares a similar structure with ViT. However, Swin Transformer adopts a hierarchical construction method resembling CNNs, which applies different ratios for down-sampling the input images, thus providing better compatibility with downstream tasks. Additionally, to enhance computational efficiency, Swin Transformer employs the methods of Windowed Multi-head Self-Attention (W-MSA) and Shifted Windowed Multi-head Self-Attention (SW-MSA) instead of the Multi-head Self-Attention (MSA) used in ViT. This approach allows the Multi-head self-attention mechanism to operate within each window and enables information interaction between windows using a shifted window technique.

In practice, Swin Transformer introduces relative position bias parameters $B$ to enhance the accuracy of attention computation.

$$\text{Attention}(Q, K, V) = \text{SoftMax}(\frac{QK^T}{\sqrt{d}} + B)V \qquad (1)$$

Where $Q$, $K$, and $V \in \mathbb{R}^{M^p \times d}$ refer to the query matrix, key matrix, and value matrix, respectively. $M^p$ represents the number of blocks in the window, and $d$ denotes the dimension of $Q$ or $K$. The introduced relative positional bias parameter $B \in \mathbb{R}^{M^p \times M^p}$ is derived from the parameterized bias matrix $\hat{B} \in \mathbb{R}^{(2M-1)\times(2M-1)}$. $M \times M$ is defined as the window size.

The multi-head attention mechanism divides the input data into multiple subspaces and assigns an independent attention weight to each subspace. In this way, it is possible to simultaneously focus on different parts of the input image to capture features at different levels of abstraction.

$$\text{MultiHead}(Q, K, V) = \text{Concat}(\text{head}_1,..., \text{head}_h)W^O \qquad (2)$$

$$\text{where head}_i = \text{Attention}(QW_i^Q, KW_i^K, VW_i^V) \qquad (3)$$

Where $W^Q$, $W^K$, and $W^V$ refer to the weight matrices used for linear transformations in $Q$, $K$, and $V$, respectively. $W^O$ is a weight matrix used for projection, aiming to integrate the attention outputs from multiple heads and map them to the final output space. Additionally, $h$ represents the number of heads in the multi-head attention mechanism, and $i$ represents the index of a head, referring to the specific head within the multi-head attention mechanism.

The process of processing and recognizing vehicle logo images is as follows: Firstly, we input a three-channel RGB image with a size of $X \in \mathbb{R}^{H \times W \times 3}$, where $H$ and $W$ represent the height and width of the image, respectively. The image $X$ is split into non-overlapping patches using the patch partition block. In our case, the patch size is set to $4 \times 4$, resulting in each patch having a feature dimension of $4 \times 4 \times 3 = 48$. Next, through a linear embedding layer, we map the feature values of each patch to an arbitrary dimension, transforming the feature dimension of each patch from 48 to $C$ (where $C$ represents the vector dimension). Then, each patch token is processed through several improved self-attention blocks, known as Swin Transformer blocks. At this point, the number of tokens is $H/4$, $W/4$, forming "Stage 1" along with the linear embedding layer. As the network deepens, we need to down-sample the image using patch merging layers to reduce the number of tokens and obtain a hierarchical representation. This process is illustrated in **Figure 2**. The first patch merging layer concatenates adjacent patches in groups of $2 \times 2$, and applies a linear layer to accurately merge corresponding positions, reducing the number of tokens to $H/8$, $W/8$, and increasing the output feature dimension from $C$ to $2C$. Together with the next Swin Transformer block, this is referred to as "Stage 2". Finally, we repeat the patch merging and Swin Transformer block process twice, referred to as "Stage 3" and "Stage 4", respectively. The number of tokens decreases to $H/16$, $W/16$ in Stage 3, and $H/32$, $W/32$ in Stage 4. Each stage generates a hierarchical structure similar to VGG (*27*) and ResNet (*3*) networks.

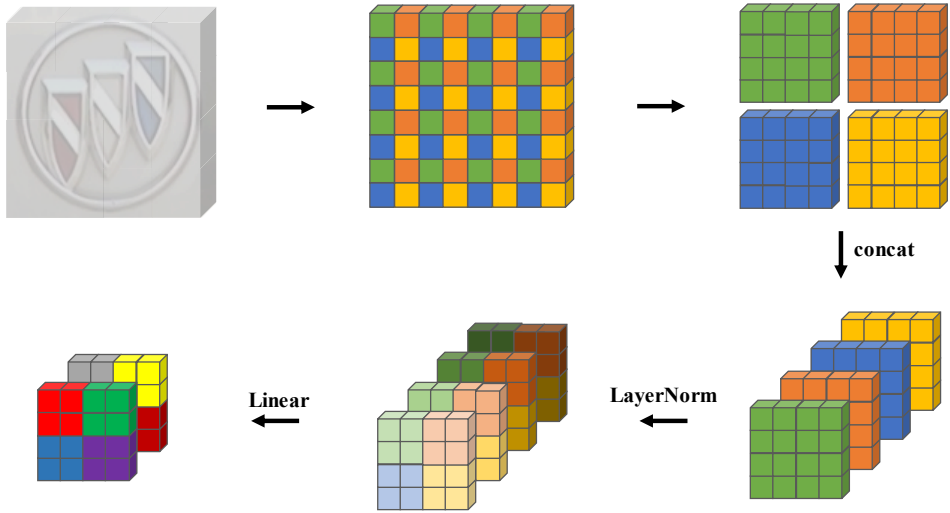

**Figure 2 Patch merging blocks**

Within the Swin Transformer blocks, MSA from ViT is replaced by W-MSA module and SW-MSA module, while the remaining structure remains unchanged. As shown in **Figure 3**. After the improved MSA module, there is a two-layer MLP with the GELU non-linear activation function. A LayerNorm (LN) layer is applied before each MSA module and MLP, and a residual connection is applied after each module. The working mechanism of the Swin Transformer hierarchical structure is as follows:

$$\hat{s}^l = \text{W-MSA}(\text{LN}(s^{l-1})) + s^{l-1} \quad (4)$$

$$s^l = \text{MLP}(\text{LN}(\hat{s}^l)) + \hat{s}^l \quad (5)$$

$$\hat{s}^{l+1} = \text{SW-MSA}(\text{LN}(s^l)) + s^l \quad (6)$$

$$s^{l+1} = \text{MLP}(\text{LN}(\hat{s}^{l+1})) + \hat{s}^{l+1} \quad (7)$$

Where $l$ represents the lth Module, and $\hat{s}^l$ and $s^l$ represent the output features of the (S)W-MSA module and the MLP module for block $l$, respectively.

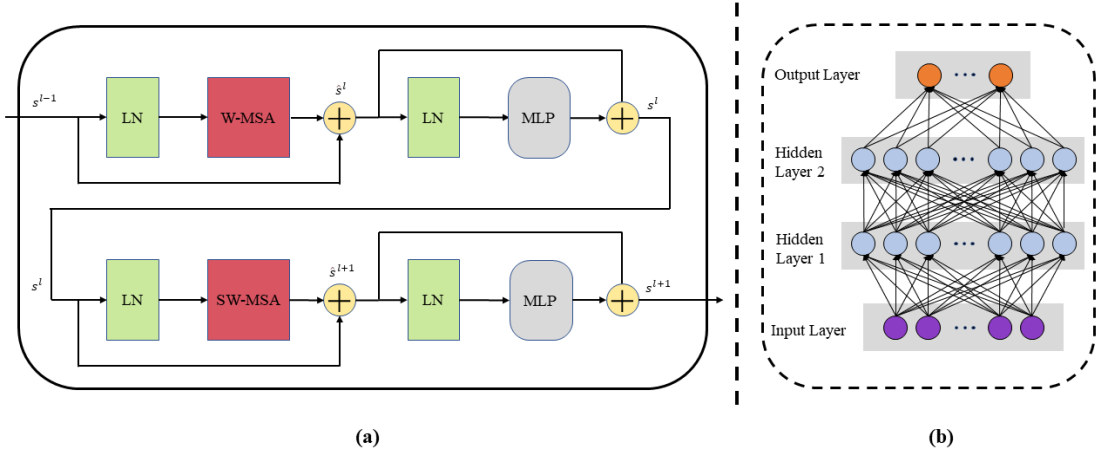

**Figure 3 Swin Transformer blocks: (a) Illustrates the internal structure of the Swin Transformer block ; (b) Shows the MLP structure within the block**

**W-MSA Block**

MSA in the ViT network primarily performs global modeling of the input image to capture rich global information. However, this global modeling approach requires integrating information from all pixels on the entire image, which undoubtedly has a negative impact on the model's computational efficiency. W-MSA in Swin Transformer effectively addresses this issue. Specifically, the Swin Transformer adopts a conventional window partitioning strategy, where the entire image is evenly partitioned into several non-overlapping windows. Then, self-attention calculations are performed separately within each non-overlapping window. Within each window, each pixel can only interact with other pixels within the same window for information exchange. W-MSA effectively reduces the redundancy in computations. The computational comparison between W-MSA and MSA is as follows:

$$\Omega(\text{MSA}) = 4hwC^2 + 2(hw)^2 C \tag{8}$$

$$\Omega(\text{W-MSA}) = 4hwC^2 + 2M^2 hwC \tag{9}$$

Where $M$ represents the size of each window, where the patch size is $h \times w$, and $C$ represents the feature dimension of the patches.

**SW-MSA Block**

W-MSA significantly reduces the computational cost by performing attention calculations within each window. However, relying solely on this approach may not capture the global information of the entire image, which is crucial for image classification and recognition. To ensure that self-attention can communicate across windows while maintaining the computational efficiency of W-MSA, Swin Transformer utilizes a shifted W-MSA method, which is alternately used between paired Swin Transformer blocks. SW-MSA approach consists of two consecutive W-MSA calculations. Firstly, a regular W-MSA calculation is performed. Then, a second W-MSA calculation is performed using a sliding window technique. Both MSA calculations incorporate information from pixels within the window as well as from other offset windows, achieving efficient modeling across windows. This enables Swin Transformer to capture both local and global information effectively.

As shown in **Figure 4**, using the regular window partition strategy of W-MSA, we divide a $8 \times 8$ feature map into $2 \times 2$ windows of size $4 \times 4$. Next, we shift all windows to the lower right by two blocks, resulting in a total of nine windows with different configurations. Subsequently, we concatenate these nine windows using a sliding window technique, forming four windows that are similar in size and structure to W-MSA. To prevent information disruption caused by window shifting, a masking mechanism is applied. Additionally, the information windows are restored to their original positions using backward cyclic shifting to extract the feature information. SW-MSA method facilitates information interaction across windows, ensuring efficient information matching and preserving the integrity of the overall image representation.

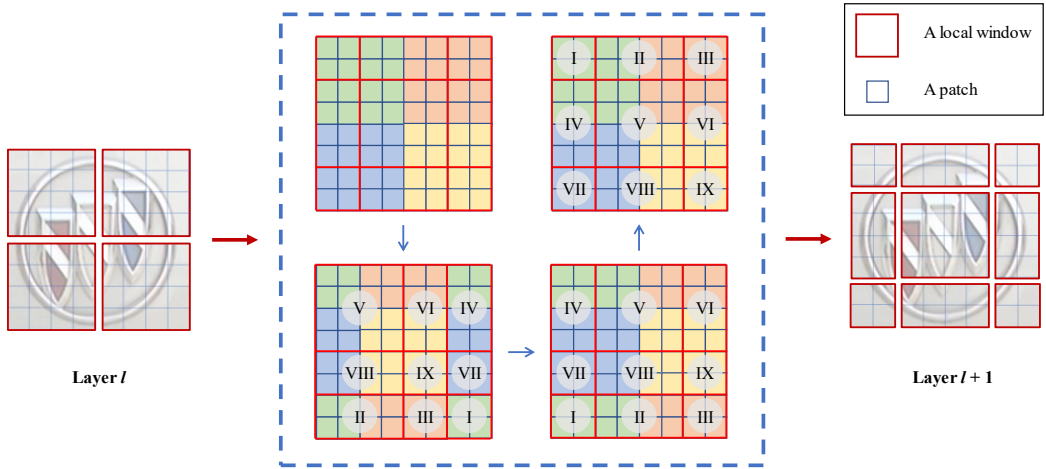

**Figure 4 Calculation method used for self-attention in shifted window partitions**

**EXPERIMENTAL RESULTS AND DISCUSSION**

The method employed in this paper was implemented using the PyTorch deep learning framework. The experiments were conducted on a machine with an NVIDIA GeForce RTX 2080 GPU and 8GB of VRAM. We conducted comprehensive experiments on three publicly available datasets: HFUT-VL (*28*), XMU (*19*), and CTGU-VLD (*29*), to evaluate the performance of the Swin Transformer method in the field of VLR. The partial sample images of these datasets are shown in **Figure 5**. The input image size was set to $224 \times 224$, and the batch size was set to 32. The initial learning rate was set to 0.0001, and a learning rate decay strategy was adopted, reducing the learning rate by a factor of 5 every 40 epochs. Data augmentation techniques such as random flipping and cropping were used to improve the model's robustness during the training process.

**Dataset**

The HFUT-VL(*28*) dataset consists of 32,000 vehicle logo images captured from a real-time highway detection system. It includes two sub-datasets, namely HFUT-VL1 and HFUT-VL2. HFUT-VL1 sub-dataset contains 16,000 vehicle logo images. There are 80 different logo categories, with each category having 200 precisely localized vehicle logo images of size $64 \times 64$. HFUT-VL2 sub-dataset has the same number of samples as HFUT-VL1, consisting of 80 logo categories as well. However, each category in HFUT-VL2 includes 200 roughly localized vehicle logo images of size $64 \times 96$.

The XMU(*19*) dataset consists of vehicle logo images from the top ten popular manufacturers, with a total of 11,500 vehicle logo images. Each vehicle logo category includes 1,150 images of size $70 \times 70$. Each image in the dataset has undergone various forms of distortion, including different lighting conditions, rotations, and noise. These distortions are applied to increase the dataset's diversity and challenge the model's robustness.

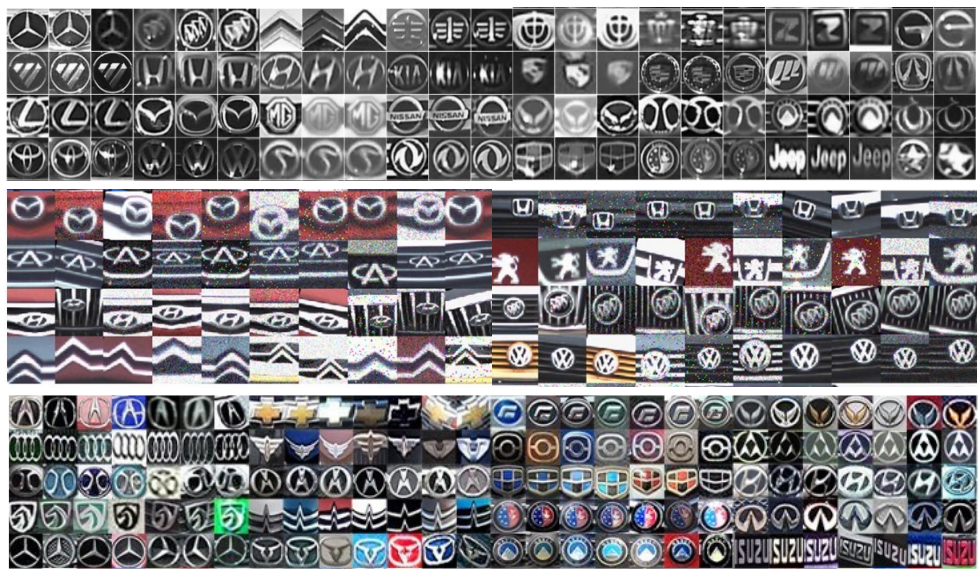

**Figure 5 Sample images from three vehicle logo datasets: HFUT-VL1, XMU, and CTGU-VLD (from top to bottom, respectively)**

The CTGU-VLD(*29*) dataset is derived from two sources: one part is captured from surveillance cameras on traffic roads, and the other part is collected from vehicle logo images extracted from the internet. The dataset contains a total of 60,000 vehicle logo images, representing 131 different logo classes. All images have been normalized to a size of $80 \times 80$. As shown in **Figure 6**, the distribution of data in CTGU-VLD is highly uneven, with some classes having as many as 1,323 images, while others have as few as 66 images. Each image has undergone data augmentation processes, such as changes in brightness, color variations, rotations, and the addition of noise, to enhance the dataset's diversity and robustness.

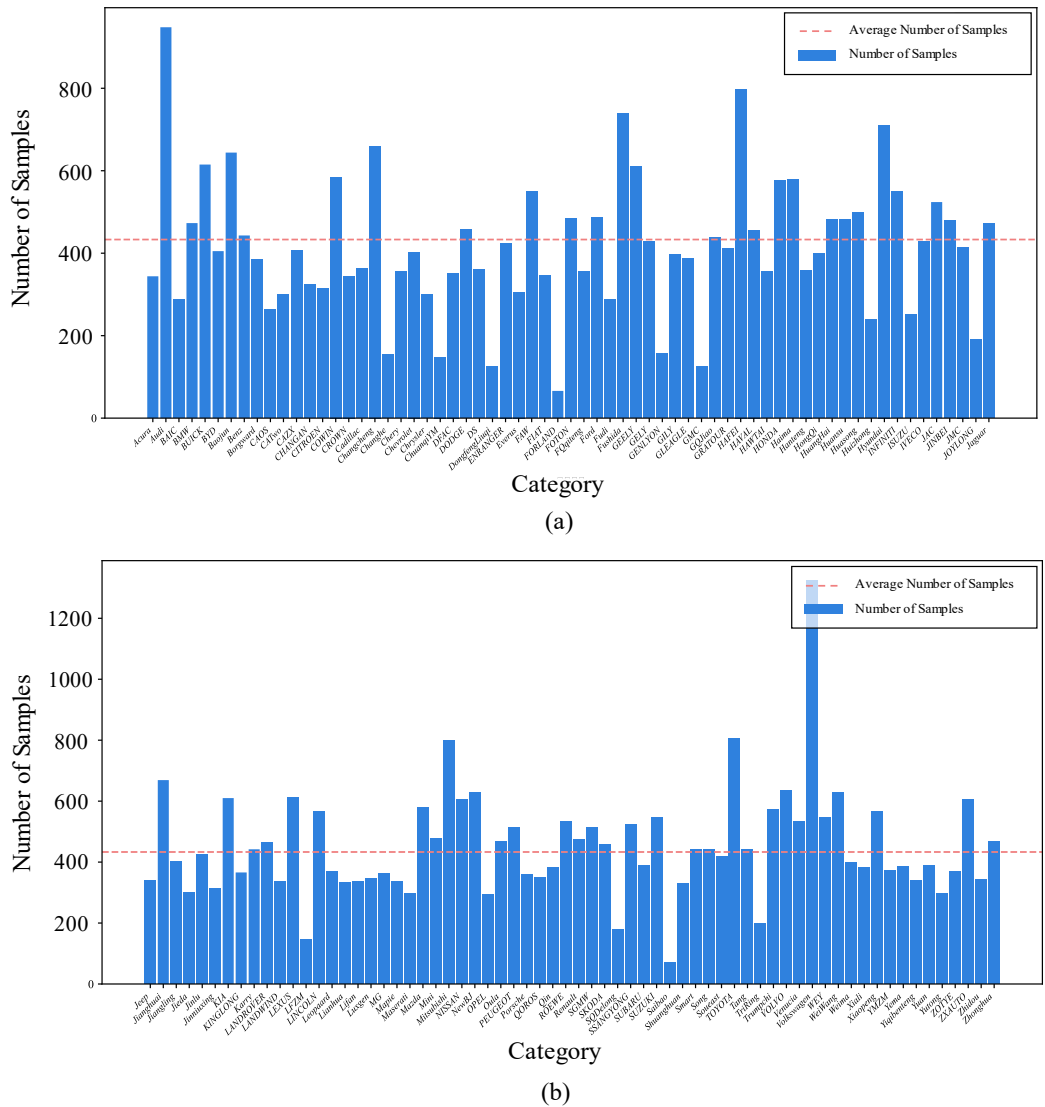

**Figure 6** Sample distribution of the CTGU-VLD vehicle logo dataset: (a) Represents the first 65 classes; (b) Represents the last 66 classes

**Parameter Evaluation**

For most classification tasks, Swin Transformer has shown good performance (*10*). However, vehicle logos belong to small objects, and in most datasets, the type, location, size, and real-world environment of vehicle logos pose significant challenges to recognition. To *Li, Zhang, and Xiao*achieve the best performance of the Swin Transformer model in VLR, we conducted experiments on the HFUT-VL vehicle logo dataset to adjust the model's parameters. The adjusted parameters include network depth, attention heads, window size, and embedding dimensions. The following summary provides specific experimental details.

*Network Depth*

Swin Transformer model consists of four stages (stage 1, stage 2, stage 3, and stage 4). In previous extensive experiments, the depth of stage 3 has been found to have the most significant impact on the performance of the Swin Transformer model, while the depths of the remaining stages are all two layers. To achieve the best performance of the model, we designed experiments to investigate the influence of different network depths in stage 3 on VLR results.

Firstly, we set the network depth to 2, which is the minimum depth, consistent with the depths of the other stages. Then, we increased the depth of the network, hoping to obtain better results. As shown in **Figure 7(a)**, we found that the performance of the model improved when the network depth was increased to 4 compared to the performance with a depth of 2. It is evident that increasing the network depth can enhance the accuracy of the Swin Transformer model.

Additionally, we further determined the optimal number of layers for the network. From **Figure 7(a)**, we observed that increasing the network depth from 4 layers to 6 layers and from 8 layers to 10 layers resulted in a decrease in accuracy. However, when the network depth was increased from 6 layers to 8 layers, the accuracy increased and reached the highest point. Although the accuracy also improved when the network depth was increased from 10 layers to 12 layers, considering that deeper networks increase model complexity, we concluded that the optimal network depth for achieving the best performance is 8 layers.

*Attention Heads*

Swin Transformer utilizes a multi-head attention mechanism to perform parallel computations on input features, capturing different levels of attention. The number of attention heads determines the parallel computations of the self-attention mechanism in each stage, thereby affecting the model's performance.

To evaluate the impact of attention heads on the model's performance, we designed four sets of attention head numbers for each of the four stages of Swin Transformer. The number of attention heads in each stage is twice the number in the previous stage. As shown in **Figure 7(b)**, as the number of attention heads increases, the accuracy gradually improves. The highest accuracy is achieved when the number of attention heads is $(3,6,12,24)$. Furthermore, increasing the number of attention heads to $(4,8,16,32)$ slightly reduces the recognition rate, but it still maintains a high level of performance. The multi-head attention mechanism is designed to capture features at different levels and directions. However, too many attention heads can rapidly increase model complexity and lead to insufficient interaction among feature information. Some heads may overly focus on local features while neglecting global features, thereby impacting the overall performance of the model. Therefore, we do not continue increasing the number of attention heads but consider $(3,6,12,24)$ as the optimal choice for attention head numbers.

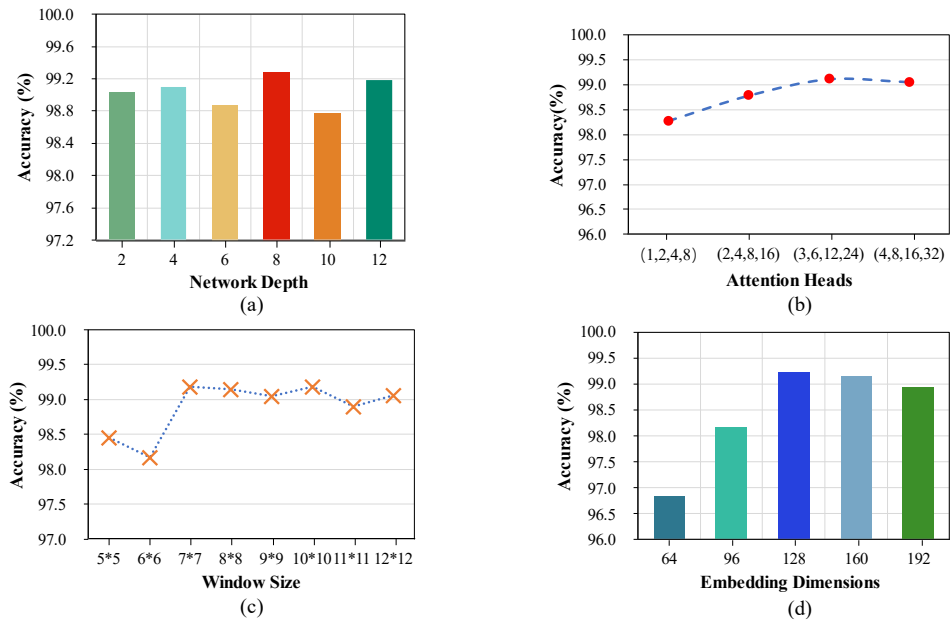

**Figure 7** Shows the parameter tuning results for the internal structure of the Swin Transformer: (a) Represents the impact of the depth of the stage 3 network on accuracy; (b) Represents the impact of the number of heads in the multi-head attention mechanism on accuracy; (c) Represents the impact of window size on accuracy; (d) Represents the impact of embedding dimension on accuracy

*Window Size*

Swin Transformer model performs multi-head self-attention calculations based on windows, effectively avoiding the massive computation brought by global self-attention while enhancing the interaction capability of information for each patch in the image. Different window sizes affect the range of interaction among each patch and its surrounding region in the self-attention mechanism. Therefore, adjusting the window size helps the model capture longer-range dependencies and global contextual information, which is beneficial for better understanding the overall shape and structure of vehicle logos in the VLR task.

We conducted experiments to evaluate the impact of window size on accuracy using eight different window sizes. From **Figure 7(c)**, we observe that increasing the window size from $5 \times 5$ to $7 \times 7$ leads to an increase in accuracy, reaching the maximum. However, when the window size is increased from $7 \times 7$ to $9 \times 9$, the accuracy decreases. When the window size is set to $10 \times 10$, the accuracy is very close to that of the $7 \times 7$ window size. Further increasing the window size results in lower accuracy compared to the 10x10 window size. Although larger windows can better capture global features, in real traffic images, vehicle logos occupy a relatively small proportion of the entire image and rely more on local details. Therefore, the $7 \times 7$ window size is more suitable for practical VLR tasks.

*Embedding Dimensions*

In addition to network depth, attention heads, and window size, the size of the embedding dimensions also affects the accuracy. In the Swin Transformer model, the embedding dimensions are denoted by "*C*," which determines the length of the feature vector at each position. On one hand, higher embedding dimensions can provide more feature dimensions, thereby increasing the model's expressive power. This means that the model can learn more rich and complex feature representations, better capturing the detailed characteristics of vehicle logos. On the other hand, lower embedding dimensions limit the model's expressive power, making it unable to effectively represent the features of input images. However, higher embedding dimensions are not always better. Higher embedding dimensions increase computation and memory overhead and also increase the risk of overfitting, which can have a negative impact on experimental results. Based on this, we analyze the experimental results to select the most suitable embedding dimensions. From **Figure 7(d)**, we observe that the model achieves the best performance when the embedding dimension is set to 128.

**Performance evaluation of different methods**

In this section, we conducted performance evaluation experiments for two groups of VLR methods. Firstly, under the condition of not loading pre-trained weights, we assessed the performance of several mainstream CNN backbone networks and ViT visual models in comparison to our method. Secondly, we loaded pre-trained weight models and evaluated their performance against the state-of-the-art VLR methods.

*Evaluation of common backbone network with CNN*

As shown in **Table 1**, we conducted performance comparisons between our algorithm and general CNN recognition models on the HFUT-VL1, XMU, and CTGU-VLD vehicle logo datasets. It is worth noting that, to obtain fair experimental results, all models were trained without using pre-trained weights.

**TABLE 1 Comparison of our method with general CNN recognition models on publicly available datasets HFUT-VL1, XMU, and CTGU-VLD. The best value for each column is highlighted in bold, and the second-best performance is indicated by underlining**

| Model | Datasets | | | | | |
|---|---|---|---|---|---|---|
| | HFUT-VL1 | | XMU | | CTGU-VLD | |
| | Accuracy (%) | Speed (images/s) | Accuracy (%) | Speed (images/s) | Accuracy (%) | Speed (images/s) |
| ResNet34(*3*) | 97.21 | **143** | 98.29 | **143** | 96.70 | 63 |
| ResNet50(*3*) | 96.88 | 111 | 97.84 | 111 | 96.49 | 53 |
| ResNet101(*3*) | 96.31 | 59 | 97.27 | 71 | 96.27 | 42 |
| MobileNet(*4*) | 94.95 | **143** | 98.34 | <u>125</u> | 92.61 | <u>125</u> |
| ShuffleNet(*5*) | 96.17 | <u>125</u> | <u>99.55</u> | 111 | 93.19 | **143** |
| EfficientNet(*6*) | 98.11 | 100 | **100** | 100 | 99.11 | **143** |
| ViT(*9*) | <u>98.84</u> | 111 | **100** | 111 | <u>99.14</u> | 111 |
| SwinT(*10*) | **99.28** | **143** | **100** | **143** | 99.17 | **143** |

For the HFUT-VL1 dataset, our method achieved the highest accuracy of 99.28% and had similar recognition times compared to the ResNet34 and EfficientNet lightweight networks. In terms of recognition accuracy, our method outperforms ViT by 0.44% and surpasses other CNN networks. Regarding the average recognition speed, our method is faster than both ViT and most CNN models. This is mainly due to the use of the sliding window-based multi-head self-attention mechanism in Swin Transformer. The XMU dataset is an open and popular dataset in the field of VLR. Recently, the XMU dataset has been used to evaluate the performance of various VLR algorithms. In the XMU dataset, our method achieved the same highest accuracy of 100% as EfficientNet and ViT. However, our method outperformed both EfficientNet and ViT in terms of average recognition speed, being 0.003s faster than EfficientNet and 0.002s faster than ViT. CTGU-VLD is a recently added vehicle logo dataset, which contains a large number of categories and exhibits an extremely imbalanced distribution. Some of the vehicle logos in this dataset are difficult even for human eyes to recognize. Although our method experiences a slight decrease in accuracy, it still demonstrates significant superiority compared to other CNN models, achieving the highest accuracy of 99.17%. Additionally, despite the highly uneven data distribution in the CTGU-VLD dataset, our method maintains its recognition speed without any degradation, remaining on par with ShuffleNet and EfficientNet.

*Evaluation with the most advanced VLR methods*

As shown in **Table 2**, we compared our method's performance with state-of-the-art VLR methods on the HFUT-VL1 and XMU vehicle logo datasets. Additionally, we loaded pre-trained weights for our method to achieve further performance improvements.

**TABLE 2 Comparison of our method with state-of-the-art VLR methods on publicly available datasets HFUT-VL1 and XMU. The best value for each column is highlighted in bold, and the second-best performance is indicated by underlining**

| Method | Method Category | Accuracy (%) | |
| --- | --- | --- | --- |
| | | HFUT-VL1 | XMU |
| Yu (*16*) | Classical | 96.30 | - |
| Lu (*17*) | Classical | 97.80 | 98.40 |
| Yu (*18*) | Classical | 98.92 | 99.98 |
| Huang (*19*) | Deep learning | - | 99.07 |
| Soon (*20*) | Deep learning | - | 99.13 |
| Chen (*21*) | Deep learning | - | **100.0** |
| Yu (*22*) | Deep learning | **99.50** | 99.80 |
| SwinT (*10*) | Deep learning | 99.28 | **100.0** |
| Swin-T (*10*) | Deep learning | 99.49 | **100.0** |

In this experiment, we compared with state-of-the-art VLR methods. We employed a transfer learning strategy by loading pre-trained weights from Swin-T to further enhance model performance. The data referenced in **Table 2** are obtained from the corresponding papers. By loading pre-trained weights, we achieved a more impressive performance for the Swin-T model. In the HFUT-VL1 dataset, our method achieved an accuracy of 99.49%. Compared to the method proposed by (*22*), although our accuracy decreased by approximately 0.01%, our method required fewer computational resources, with a complexity comparable to ResNet50. Even without loading pre-trained weights, Swin Transformer still performed on par with or even surpassed the majority of advanced VLR methods. In the XMU dataset, our method exhibited the highest accuracy, comparable to (*21*). It is noteworthy that Swin-T reached the highest accuracy in a shorter training time.

In summary, our method achieved the highest accuracy on HFUT-VL1, XMU, and CTGU-VLD, with accuracies of 99.28%, 100%, and 99.17%, respectively, without loading pre-trained weights. The recognition speed is comparable to lightweight CNN models. Compared to other CNN models, our approach demonstrates superiority. Additionally, loading pre-trained weights further enhances performance. Swin-T delivers remarkable results in a short time, achieving accuracies of 99.49% and 100% on HFUT-VL1 and XMU, respectively, with complexity similar to ResNet, on par with or even surpassing state-of-the-art methods.

**CONCLUSIONS**

In this paper, we achieved the optimal performance of the model in VLR by fine-tuning the internal structure parameters of the Swin Transformer. We obtained the highest accuracy rates of 99.28%, 100%, and 99.17% on the HFUT-VL1, XMU, and CTGU-VLD datasets, respectively. Specifically, we employed the basic structure of the Swin Transformer, which consists of multiple levels of Transformer modules, with each module composed of several attention mechanisms and fully connected layers. This hierarchical structure allows the model to model input features at different scales, thereby extracting more diverse feature representations. Additionally, we accomplished non-overlapping window interaction using the shifted window technique. Through experimental comparisons, we found that Swin Transformer, as a novel universal visual backbone network, can extract more informative features than CNN. Moreover, with the simple approach of loading pre-trained weights, we achieved competitive performance compared to state-of-the-art VLR methods.

In conclusion, our research successfully leveraged the Swin Transformer for real-time VLR and achieved outstanding accuracy on multiple public datasets. Future studies can explore the integration of VLR with comprehensive recognition of vehicle models, colors, and other features to enhance overall vehicle recognition performance. We believe that, with further improvements and optimizations, the Swin Transformer will demonstrate even broader applications in practical scenarios.

**ACKNOWLEDGMENTS**

This work is supported by China NSFC Program under Grant NO. 61603257.

**AUTHOR CONTRIBUTIONS**

The authors confirm contribution to the paper as follows: study conception and design: Z. D., X. J.; data collection: L. Y., Z. D.; analysis and interpretation of results: L. Y., Z. D., X. J.; draft manuscript preparation: L. Y. All authors reviewed the results and approved the final version of the manuscript.

*Li, Zhang, and Xiao***REFERENCES**

1. Huang, G.-L., Zaslavsky, A., Loke, S. W., Abkenar, A., Medvedev, A., & Hassani, A. (2023). Context-Aware Machine Learning for Intelligent Transportation Systems: A Survey. IEEE Transactions on Intelligent Transportation Systems, 24(1), 17–36.

2. Jiang, Y., Jiang, F., Luo, H., Lin, H., Yao, J., Liu, J., and Ren, J. (2023). An Efficient and Unified Recognition Method for Multiple License Plates in Unconstrained Scenarios. IEEE Transactions on Intelligent Transportation Systems, 24(5), 5376–5389.

3. He, K., Zhang, X., Ren, S., and Sun, J. (2016). Deep Residual Learning for Image Recognition. Proceedings of the IEEE Conference on Computer Vision and Pattern Recognition (CVPR), 770–778.

4. Sandler, M., Howard, A., Zhu, M., Zhmoginov, A., and Chen, L.-C. (2018). MobileNetV2: Inverted Residuals and Linear Bottlenecks. Proceedings of the IEEE Conference on Computer Vision and Pattern Recognition (CVPR), 4510–4520.

5. Zhang, X., Zhou, X., Lin, M., and Sun, J. (2018). ShuffleNet: An Extremely Efficient Convolutional Neural Network for Mobile Devices. Proceedings of the IEEE Conference on Computer Vision and Pattern Recognition (CVPR), 6848–6856.

6. Tan, M., and Le, Q. (2019). EfficientNet: Rethinking Model Scaling for Convolutional Neural Networks. Proceedings of the 36th International Conference on Machine Learning(ICML), 6105–6114.

7. Raghu, M., Unterthiner, T., Kornblith, S., Zhang, C., and Dosovitskiy, A. (2021). Do Vision Transformers See Like Convolutional Neural Networks? Advances in Neural Information Processing Systems, 34, 12116–12128.

8. Vaswani, A., Shazeer, N., Parmar, N., Uszkoreit, J., Jones, L., Gomez, A. N., Kaiser, L., and Polosukhin, I. (2017). Attention Is All You Need. Part of Advances in Neural Information Processing Systems 30 (NIPS), 5998–6008.

9. Dosovitskiy, A., Beyer, L., Kolesnikov, A., Weissenborn, D., Zhai, X., Unterthiner, T., Dehghani, M., Minderer, M., Heigold, G., Gelly, S., Uszkoreit, J., and Houlsby, N. (2021). An Image is Worth 16x16 Words: Transformers for Image Recognition at Scale. International Conference on Learning Representations (ICLR).

10. Liu, Z., Lin, Y., Cao, Y., Hu, H., Wei, Y., Zhang, Z., Lin, S., & Guo, B. (2021). Swin Transformer: Hierarchical Vision Transformer using Shifted Windows. 2021 IEEE/CVF International Conference on Computer Vision (ICCV), 9992–10002.16

11. Liang, J., Cao, J., Sun, G., Zhang, K., Van Gool, L., & Timofte, R. (2021). SwinIR: Image Restoration Using Swin Transformer. 2021 IEEE/CVF International Conference on Computer Vision Workshops (ICCVW), 1833–1844.

12. Liu, Z., Hu, H., Lin, Y., Yao, Z., Xie, Z., Wei, Y., Ning, J., Cao, Y., Zhang, Z., Dong, L., Wei, F., & Guo, B. (2022). Swin Transformer V2: Scaling Up Capacity and Resolution. 2022 IEEE/CVF Conference on Computer Vision and Pattern Recognition (CVPR), 11999–12009.

13. Liu, Z., Ning, J., Cao, Y., Wei, Y., Zhang, Z., Lin, S., & Hu, H. (2022). Video Swin Transformer. 2022 IEEE/CVF Conference on Computer Vision and Pattern Recognition (CVPR), 3192–3201.

14. Psyllos, A., Anagnostopoulos, C.-N., and Kayafas, E. (2012). M-SIFT: A new method for Vehicle Logo Recognition. 2012 IEEE International Conference on Vehicular Electronics and Safety (ICVES), 261–266.

15. Yu, S., Zheng, S., Yang, H., and Liang, L. (2013). Vehicle logo recognition based on Bag-of-Words. 2013 10th IEEE International Conference on Advanced Video and Signal Based Surveillance (AVSS), 353–358.

16. Yu, Y., Wang, J., Lu, J., Xie, Y., and Nie, Z. (2018). Vehicle logo recognition based on overlapping enhanced patterns of oriented edge magnitudes. Computers & Electrical Engineering, 71, 273–283.

17. Lu, L., and Huang, H. (2019). A Hierarchical Scheme for Vehicle Make and Model Recognition From Frontal Images of Vehicles. IEEE Transactions on Intelligent Transportation Systems, 20(5), 1774–1786.

18. Yu, Y., Li, H., Wang, J., Min, H., Jia, W., Yu, J., and Chen, C. (2021). A Multilayer Pyramid Network Based on Learning for Vehicle Logo Recognition. IEEE Transactions on Intelligent Transportation Systems, 22(5), 3123–3134.

19. Huang, Y., Wu, R., Sun, Y., Wang, W., and Ding, X. (2015). Vehicle Logo Recognition System Based on Convolutional Neural Networks With a Pretraining Strategy. IEEE Transactions on Intelligent Transportation Systems, 16(4), 1951–1960.

20. Soon, F. C., Khaw, H. Y., Chuah, J. H., and Kanesan, J. (2019). Vehicle logo recognition using whitening transformation and deep learning. Signal, Image and Video Processing, 13(1), 111–119.

21. Chen, R., Mihaylova, L., Zhu, H., and Bouaynaya, N. C. (2020). A Deep Learning Framework for Joint Image Restoration and Recognition. Circuits, Systems, and Signal Processing, 39(3), 1561–1580.

22. Yu, Y., Guan, H., Li, D., and Yu, C. (2021). A Cascaded Deep Convolutional Network for Vehicle Logo Recognition From Frontal and Rear Images of Vehicles. IEEE Transactions on Intelligent Transportation Systems, 22(2), 758–771.

23. Wang, D., Al-Rubaie, A., Alsarkal, Y. I., Stincic, S., and Davies, J. (2021). Cost effective and Accurate Vehicle Make/Model Recognition method Using YoloV5. 2021 International Conference on Smart Applications, Communications and Networking (SmartNets), 1–4.

24. Agarwal, A., Shinde, S., Mohite, S., and Jadhav, S. (2022). Vehicle Characteristic Recognition by Appearance: Computer Vision Methods for Vehicle Make, Color, and License Plate Classification. 2022 IEEE Pune Section International Conference (PuneCon), 1–6.

25. Amirkhani, A., and Barshooi, A. H. (2023). DeepCar 5.0: Vehicle Make and Model Recognition Under Challenging Conditions. IEEE Transactions on Intelligent Transportation Systems, 24(1), 541–553.

26. Shi, X., Ma, S., Shen, Y., Yang, Y., and Tan, Z. (2023). Vehicle logo detection using an IoAverage loss on dataset VLD100K-61. EURASIP Journal on Image and Video Processing, 2023(1), 4.

27. Simonyan, K., and Zisserman, A. (2015). Very Deep Convolutional Networks for Large-Scale Image Recognition (arXiv:1409.1556). arXiv.

28. Yu, Y., Wang, J., Lu, J., Xie, Y., and Nie, Z. (2018). Vehicle logo recognition based on overlapping enhanced patterns of oriented edge magnitudes. Computers & Electrical Engineering, 71, 273–283.

29. Li, Ne., Xu, G., Lei, B., Ma, G., and Shi, Y. (2022). Logo recognition algorithm for vehicles on traffic road. Journal of Computer Applications, 2022, 42(3): 810-817.